\documentclass[10pt,conference]{IEEEtran}
\IEEEoverridecommandlockouts
\usepackage{cite}
\usepackage{amsmath,amssymb,amsfonts}
\usepackage{algorithmic}
\usepackage{graphicx}
\usepackage{textcomp}
\usepackage{float}
\usepackage{xcolor}
\usepackage{algorithm}
\usepackage{algorithmic}
\usepackage{multirow}
\usepackage{dblfloatfix}
\usepackage{placeins}
\usepackage{hyperref}

\def\BibTeX{{\rm B\kern-.05em{\sc i\kern-.025em b}\kern-.08em
    T\kern-.1667em\lower.7ex\hbox{E}\kern-.125emX}}
\begin{document}

\title{Instance Hardness–Based Relevance for Imbalanced Regression\\

\thanks{}
}

\author{
\IEEEauthorblockN{Vitor M. Leitao$^{\dagger}$, Juscimara G. Avelino$^{\dagger}$,
George D. C. Cavalcanti$^{\dagger}$,
Rafael M. O. Cruz$^{\ddagger}$}

\IEEEauthorblockA{$^{\dagger}$Centro de Informática, Universidade Federal de Pernambuco, Recife, Brazil}

\IEEEauthorblockA{$^{\ddagger}$École de Technologie Supérieure, University of Quebec, Montreal, Canada}

\IEEEauthorblockA{\texttt{\{vml2, jga2, gdcc\}@cin.ufpe.br, rafael.menelau-cruz@etsmtl.ca}}
}


\maketitle

\begin{abstract}
Imbalanced regression problems arise when the target variable has an asymmetric distribution, resulting in underrepresented value ranges in the dataset. Traditional approaches for identifying rare instances rely on a relevance function that assigns higher importance to specific regions of the target distribution. However, the effectiveness of imbalance-aware learning methods depends strongly on how relevance is defined. In more complex scenarios, such as bimodal distributions, traditional relevance functions struggle to capture rarity, as they assign fixed relevance values based solely on target values, thereby compromising the distinction between truly rare and normal instances. To address these limitations, this study proposes an Instance Hardness-based relevance function (InHaR) for identifying rare instances in regression problems. Unlike traditional relevance functions, the proposed approach incorporates learning difficulty, allowing rarity to be inferred not only from the target distribution but also from the difficulty of instances for the learning algorithm. This property is particularly important in bimodal scenarios, where rarity cannot be accurately inferred from target values alone. Experimental results demonstrate that the InHaR correctly identifies rare regions under bimodal distributions and, when used to guide resampling strategies such as Random Oversampling (RO) and Gaussian Noise (GN), leads to significant improvements in predictive performance compared to traditional relevance-based approaches. 
The code, dataset, and further details about the proposed method are publicly available at \url{https://github.com/VitorLeitao/instance-hardness-Imbalanced-regression}.
\end{abstract}


\section{Introduction}

In machine learning, imbalanced problems occur when the data distribution is asymmetric, so that certain regions of the space of interest contain a significantly smaller number of instances than other, better-represented regions. This scenario has been widely studied in classification tasks \cite{haixiang2017learning, krawczyk2016learning, johnson2019survey}, but it also arises in regression problems \cite{branco2016ubl}, where the target variable exhibits non-uniform distributions characterized by the presence of rare or hard-to-model values \cite{kowatschImbalanceRegression}. Imbalanced regression can negatively impact the performance of predictive models, as they tend to favor denser regions of the distribution at the expense of less frequent regions \cite{avelino2024resampling}.

\begin{figure}[h]
    \centering
    \includegraphics[width=\columnwidth]{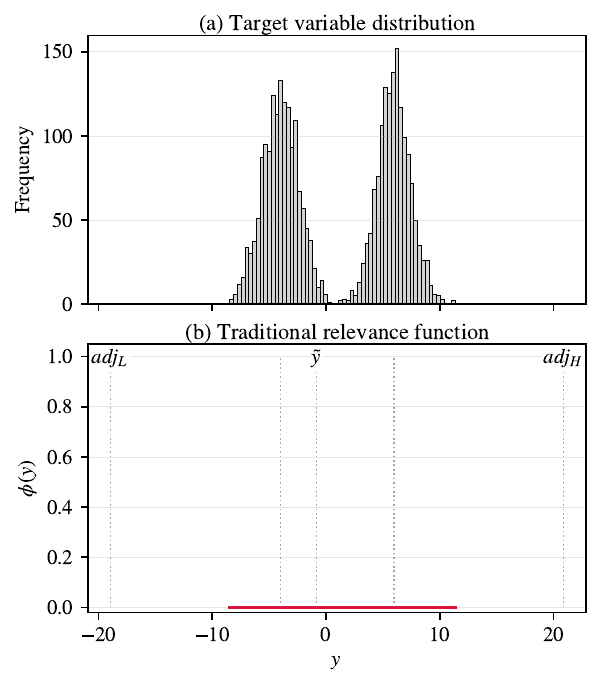}
    \caption{Bimodal target distribution and the corresponding traditional relevance function.
    (a) Histogram of the target variable.
    (b) Tukey-based relevance function $\phi(y)$~\cite{ribeiro2011utility}, whose control points are derived from boxplot statistics. The function is anchored at the median $\tilde{y}$ and at the adjacent lower and upper limits, defined as $adj_L = Q1 - 1.5 \cdot IQR$ and $adj_H = Q3 + 1.5 \cdot IQR$, where $Q1$ and $Q3$ denote the first and third quartiles and $IQR = Q3 - Q1$. Values outside these limits are assigned higher relevance.}
    \label{fig:bimodal}
\end{figure}

The identification of rare instances is a fundamental step and has traditionally been addressed in the literature through a relevance function\footnote{Throughout this work, the term \emph{traditional relevance function} refers to the relevance function proposed by Ribeiro (2011)~\cite{ribeiro2011utility}.}, which classifies instances as rare or normal based on the target variable distribution. In this context, rare instances correspond to observations in extreme or low-density regions of the distribution, whereas normal instances comprise the majority of samples associated with frequent target values. This identification is required for applying data resampling strategies such as Random Oversampling (RO) and Gaussian Noise (GN)~\cite{branco2019pre}. However, despite its widespread use, the relevance function proposed by Ribeiro (2011)~\cite{ribeiro2011utility} presents important limitations, particularly in bimodal distributions. In such scenarios, relevance is assigned based on global statistical thresholds derived from the target values, which can lead to an inadequate characterization of rarity. As a consequence, the function fails to distinguish rare from normal samples located in low-density regions between modes, assigning uniformly low or near-zero relevance to these instances and effectively treating them as non-relevant~\cite{stocksieker2025comprehensive, inkim2025distance}. Figure~\ref{fig:bimodal} illustrates this behavior.

In light of this limitation, this work proposes an Instance Hardness-based relevance function (InHaR) for regression problems, grounded in the Instance Hardness (IH) concept~\cite{paiva2022instancehardness}, which quantifies how difficult an instance is for a learning algorithm to correctly predict. The motivation for this approach is supported by findings in the literature indicating that rare cases are often associated with higher prediction difficulty and tend to be poorly modeled by standard learning algorithms~\cite{ribeiro2011utility, kowatschImbalanceRegression}. Leveraging this relationship, the proposed function defines relevance directly in terms of learning difficulty, overcoming limitations of traditional relevance function, particularly their inability to distinguish different degrees of rarity in complex target distributions such as bimodal scenarios.

Unlike relevance functions based exclusively on statistical rarity, Instance Hardness captures the behavior of the learning process itself, allowing instances to be considered relevant even when they are frequent in the data distribution but systematically mispredicted. By moving beyond rarity as a purely distributional concept, this perspective explicitly considers overall predictive performance, leading to a more nuanced characterization of instance importance. As a result, it enables the identification of challenging regions overlooked by traditional relevance functions and allows learning methods to focus on instances that genuinely hinder model performance.

The evaluation of the proposed approach was conducted from three main perspectives: (i) the analysis of the correlation between Instance Hardness (IH) and the traditional relevance function ($\phi$); (ii) the assessment of the InHaR function in scenarios with bimodal distributions; and (iii) the use of the InHaR in resampling strategies. In the first perspective, we investigated whether there is a significant correlation between the concepts of rarity and instance difficulty, that is, whether instances considered rare, associated with high values of $\phi$, also correspond to difficult instances. In the second perspective, the effectiveness of the InHaR function in correctly identifying different degrees of rarity in regressions with bimodal distributions was validated, highlighting its ability to overcome the limitations of the relevance function~\cite{ribeiro2011utility} in these scenarios. Finally, in the third perspective, the use of the InHaR as a criterion in resampling methods was analyzed, replacing relevance functions in the definition of rare and normal instances. The results show that, in approximately 70\% of the tests performed, the proposed approach achieved superior performance compared to the traditional approach.


\section{Related Work}

Imbalance in regression problems has been less explored than in classification, but has recently attracted increasing attention in the literature. This setting arises when certain regions of the target variable distribution are underrepresented, which can compromise a model’s ability to accurately predict rare values~\cite{kowatschImbalanceRegression}. Existing approaches to address imbalanced regression are commonly grouped into three categories: regression models, modifications to the learning process, and evaluation metrics~\cite{avelino2024resampling}. While both individual and ensemble regressors are widely used, their performance under imbalance is often improved by altering the learning process, particularly through preprocessing strategies based on resampling or cost-sensitive learning~\cite{ribeiro2011utility, branco2019pre}. Commonly used resampling techniques include Random Oversampling~\cite{branco2019pre}, Gaussian Noise~\cite{branco2016ubl}, SmoteR~\cite{torgo2013smote}, and its extensions such as SMOGN and SmoteR-Geometric~\cite{branco2017smogn, camacho2022geometric}. Finally, traditional global metrics such as MSE may fail to reflect performance in underrepresented regions of the target space, motivating the proposal of evaluation measures specifically tailored to imbalanced regression, including the Squared Error Relevance Area (SERA)~\cite{ribeiro2020imbalanced}.

In imbalanced regression, resampling strategies commonly rely on a relevance function, originally proposed in~\cite{ribeiro2011utility}, which is used to identify rare instances based on the target variable. The relevance function maps each target value $y \in Y$ to a relevance score $\phi(y) \in [0,1]$, where higher values indicate greater importance or rarity. This mapping is typically constructed using Piecewise Cubic Hermite Interpolating Polynomials (pchip) over a set of control points, defined either through domain knowledge or automatically~\cite{dougherty1989nonnegativity, ribeiro2020imbalanced}.

In its automatic formulation, the control points are derived from Tukey’s boxplot statistics~\cite{tukey1970exploratory}. Specifically, the relevance function is anchored at the median of the target variable distribution and at the adjacent lower and upper limits, defined as $adj_L = Q1 - 1.5 \cdot IQR$ and $adj_H = Q3 + 1.5 \cdot IQR$, where $Q1$ and $Q3$ are the first and third quartiles and $IQR = Q3 - Q1$. Target values outside these adjacent limits are assigned higher relevance scores, reflecting their lower frequency in the distribution. Given a user-defined relevance threshold $t_R$, instances are classified as rare or normal, guiding resampling and evaluation procedures.

Despite its widespread adoption, recent studies have highlighted limitations of the relevance function, particularly in multimodal distributions. In such scenarios, distinct regions of the target space may receive similar relevance scores, leading to misclassification of rare and normal instances across different modes~\cite{stocksieker2025comprehensive}. Distance-based alternatives, such as the Distance-Based Relevance Function~\cite{inkim2025distance}, have been proposed to alleviate this issue by incorporating local density information. Nevertheless, these approaches remain fundamentally dependent on the marginal distribution of the target variable, disregarding the attribute space and the learning difficulty associated with individual instances.

Given this scenario, a gap in the literature exists regarding rarity criteria that do not rely solely on the statistical distribution of the target variable. In this context, this work introduces a relevance criterion grounded in the concept of Instance Hardness~\cite{torquette2024instancehardness}, which quantifies the difficulty of predicting individual instances. The proposed approach does not assume that statistical rarity and learning difficulty are equivalent; instead, it defines relevance directly in terms of predictive difficulty, allowing instances that are hard to learn to be treated as rare regardless of their frequency in the target distribution. By shifting the definition of relevance from distributional properties to learning behavior, this work extends relevance-based resampling methods and offers an alternative perspective for addressing imbalanced regression.

\section{THE PROPOSED RELEVANCE FUNCTION}

The Algorithm ~\ref{alg:ih_threshold} presents the complete procedure for the proposed relevance function, called InHaR. The method receives as input a dataset $D$ and a relevance threshold $\tau$, and outputs two disjoint sets containing rare ($D_R$) and normal ($D_N$) instances. The core idea is to define rarity based on predictive difficulty, quantified through the Instance Hardness (IH) measure.

\begin{algorithm}[H]
\caption{Instance Hardness-based relevance function (InHaR)}
\label{alg:ih_threshold}
\begin{algorithmic}[1]
\REQUIRE Dataset $D = \{(\mathbf{x}_i, y_i)\}_{i=1}^{n}$, threshold $\tau$
\ENSURE Set of rare instances $D_R$, set of normal instances $D_N$

\STATE Compute the Instance Hardness values for all instances:
\STATE \hspace{0.5cm} $IH \leftarrow \texttt{instance\_hardness}(D)$

\STATE Initialize $D_R \leftarrow \emptyset$, $D_N \leftarrow \emptyset$

\FOR{$i = 1$ to $n$}
    \IF{$IH_i \geq \tau$}
        \STATE $D_R \leftarrow D_R \cup \{(\mathbf{x}_i, y_i)\}$
    \ELSE
        \STATE $D_N \leftarrow D_N \cup \{(\mathbf{x}_i, y_i)\}$
    \ENDIF
\ENDFOR

\RETURN $D_R, D_N$
\end{algorithmic}
\end{algorithm}

The first step of the algorithm consists of computing the Instance Hardness values for all instances in the dataset. In this work, IH is defined according to the formulation proposed for regression tasks in~\cite{torquette2024instancehardness}, as shown in Equation~\ref{eq:IH}. Given an instance $(x_i, y_i)$, its difficulty is estimated from the prediction errors produced by a set of regressors $\mathcal{L}$, where $h_j(x_i)$ denotes the output of regressor $j$ for instance $x_i$. The normalization term is defined as $\gamma = \frac{1}{n} \sum_i y_i^2$, ensuring scale invariance across different datasets.

\begin{equation}
IH_{\mathcal{L}}(x_i, y_i)
=
1
-
\frac{1}{|\mathcal{L}|}
\sum_{j=1}^{|\mathcal{L}|}
\exp\left(
-
\frac{d\big(y_i, h_j(x_i)\big)}{\gamma}
\right)
\label{eq:IH}
\end{equation}

As illustrated in Algorithm~\ref{alg:ih_threshold}, the computed IH values are then used as a relevance criterion. Since IH assumes continuous values in the interval $[0,1]$, higher values indicate instances that are more difficult to predict and, consequently, potentially more relevant in imbalanced regression scenarios. To separate rare and normal instances, a relative threshold $\tau$ is defined over the IH distribution. Instances with $IH \geq \tau$ are assigned to the rare set $D_R$, while those with $IH < \tau$ are classified as normal and assigned to $D_N$.

By relying on predictive difficulty rather than exclusively on the marginal distribution of the target variable, the proposed relevance function incorporates information from the attribute space and the behavior of learning models. This allows the InHaR to be seamlessly integrated into data resampling techniques such as Random Oversampling (RO) and Gaussian Noise (GN), replacing traditional relevance functions while preserving their general workflow. The specific value of the threshold $\tau$ is treated as a method parameter and is defined experimentally, as detailed in the methodology section.

\section{EXPERIMENTAL PROTOCOL}

\noindent \textbf{Datasets.} To evaluate the behavior of the resampling methods guided by the InHaR in different imbalanced regression scenarios, we conducted a comprehensive experimental investigation involving 29 datasets. The main information about the datasets used is available in Table~\ref{tab:datasets_summary}, where instances whose relevance function $\phi$~\cite{torgo2009utility} value is greater than 0.7 are considered rare.

\begin{table*}[t]
\centering
\caption{Summary of the 29 datasets used, including the number of instances ($N$), number of attributes ($p$), and the percentage of rare cases ($\%r$).}
\label{tab:datasets_summary}
\scriptsize
\setlength{\tabcolsep}{6pt}

\begin{tabular}{lccc|lccc|lccc}
\hline
\textbf{Dataset} & \textbf{$N$} & \textbf{$p$} & \textbf{$\%r$} &
\textbf{Dataset} & \textbf{$N$} & \textbf{$p$} & \textbf{$\%r$} &
\textbf{Dataset} & \textbf{$N$} & \textbf{$p$} & \textbf{$\%r$} \\
\hline

a1 & 198 & 11 & 14.14 & acceleration & 1732 & 14 & 5.14 & cpu\_small & 8192 & 12 & 8.70 \\
a2 & 198 & 11 & 11.11 & airfoild & 1503 & 5 & 10.71 & debutanizer & 2394 & 7 & 10.03 \\
a3 & 198 & 11 & 16.16 & analcatdata\_apnea3 & 450 & 11 & 22.89 & fuel\_consumption\_country & 1764 & 37 & 9.30 \\
a7 & 198 & 11 & 13.64 & available\_power & 1802 & 15 & 8.71 & forestFires & 517 & 12 & 15.28 \\
abalone & 4177 & 8 & 16.26 & boston & 506 & 13 & 17.98 & heat & 7400 & 11 & 8.97 \\
cocomo\_numeric & 60 & 56 & 16.67 & compactiv & 8192 & 21 & 8.70 & kdd\_coil\_1 & 316 & 18 & 10.76 \\
concreteStrength & 1030 & 8 & 5.34 & lungcancer\_shedden & 442 & 24 & 5.66 & maximal\_torque & 1802 & 32 & 7.16 \\
meta & 528 & 65 & 20.45 & mortgage & 1049 & 15 & 10.10 & pdgfr & 79 & 320 & 12.66 \\
sensory & 576 & 11 & 11.98 & space\_ga & 3107 & 6 & 5.57 & treasury & 1049 & 15 & 10.39 \\
triazines & 186 & 60 & 10.75 & wine & 6497 & 11 & 23.44 & & & & \\

\hline
\end{tabular}
\end{table*}

\noindent \textbf{Regression Models.} 
To assess the effect of resampling strategies guided by different relevance criteria across diverse learning paradigms, we considered a set of widely adopted regression models with distinct inductive biases: Random Forest Regressor (RF), Bagging Regressor (BG), XGBRegressor (XGB), Support Vector Regressor (SVR), and Multilayer Perceptron Regressor (MLP)\footnote{RF, BG, SVR, and MLP were implemented using the \texttt{scikit-learn} library (version 1.6.1). XGB was implemented using the \texttt{xgboost} library (version 3.1.3).}. These models were selected because they are commonly used in imbalanced regression studies~(\cite{avelino2024resampling,inkim2025distance}) and represent complementary families of learners, including ensemble-based methods, kernel-based models, and neural networks.

\noindent \textbf{Instance Hardness.}
In the experimental evaluation, Instance Hardness (IH) was used as the sole criterion to estimate instance relevance. Following the definition introduced in Section~3, IH was instantiated using error-based measures computed from the regression models defined in the \textit{Regression Models} subsection. The resulting hardness values were normalized to the range $[0,1]$, where higher values indicate greater prediction difficulty. A fixed threshold ($\tau = 0.7$) was adopted to distinguish rare from normal instances, and these IH scores were then used to guide the resampling process.

\noindent \textbf{Metrics.} To evaluate model performance across datasets, we employed two traditional regression metrics: Mean Absolute Error (MAE) and Mean Squared Error (MSE). These metrics provide a general assessment of predictive accuracy and are independent of any relevance definition. Metrics specifically designed for imbalanced regression, such as SERA~\cite{ribeiro2020imbalanced} or adaptations of the F1-score~\cite{torgo2009precision}, were not adopted as primary evaluation criteria, as they depend explicitly on a predefined relevance function. Relying on evaluation measures that directly incorporate a relevance function could introduce bias into the assessment, particularly when the relevance definition itself is under analysis, potentially favoring methods aligned with that definition rather than reflecting intrinsic predictive performance.

Since this work proposes a new relevance criterion based on Instance Hardness, relying on evaluation measures that directly incorporate a relevance function could introduce bias in favor of the proposed approach. Therefore, MAE and MSE were used as relevance-independent metrics to ensure a neutral comparison among methods. In addition to global performance, MAE and MSE were computed over subsets of instances identified as rare using two relevance criteria: the traditional relevance function and the proposed InHaR. This analysis allows us to assess model behavior across distinct definitions of rarity while maintaining a consistent, unbiased evaluation protocol.



\noindent \textbf{Evaluation Methods.}
For the experiments comparing traditional resampling approaches with the InHaR, we employed \textit{repeated $k$-fold} cross-validation. Specifically, a configuration with 5 folds and 2 repetitions (i.e., $2\times5$ cross-validation) was adopted.

\noindent \textbf{Resampling Methods.} To address data imbalance, two classical resampling methods were used: Random Oversampling (RO) and Introduction of Gaussian Noise (GN). RO consists of the random replication of rare instances, while GN generates new synthetic instances by adding Gaussian noise to the attributes of rare instances, while simultaneously removing part of the most frequent instances. These two strategies were chosen because they presented the best empirical performance in the comparative study conducted by Avelino et al. \cite{avelino2024resampling}, making them suitable references for evaluating the proposed method.

\section{RESULTS AND DISCUSSION}

This section presents and discusses the experimental results obtained with the proposed approach. The evaluation is conducted from three complementary perspectives: 
(i) the relationship between Instance Hardness (IH) and the traditional relevance function ($\phi$); 
(ii) the behavior of the InHaR in datasets with bimodal target distributions; and 
(iii) the impact of InHaR on resampling strategies and the predictive performance of regression models.

\subsection{Correlation between InHaR and $\phi$}
Initially, an analysis was conducted to investigate the relationship between the traditional relevance function $\phi$~\cite{torgo2009utility} and the proposed InHaR, assessing whether instances deemed rare by $\phi$ also correspond to difficult-to-predict instances. For this purpose, the Pearson correlation coefficient was computed between the relevance values and the corresponding Instance Hardness scores. This analysis is not intended to validate equivalence between the two criteria, but rather to assess the extent to which they capture overlapping or complementary information.

\begin{figure}[h]
\includegraphics[width=0.49\textwidth]{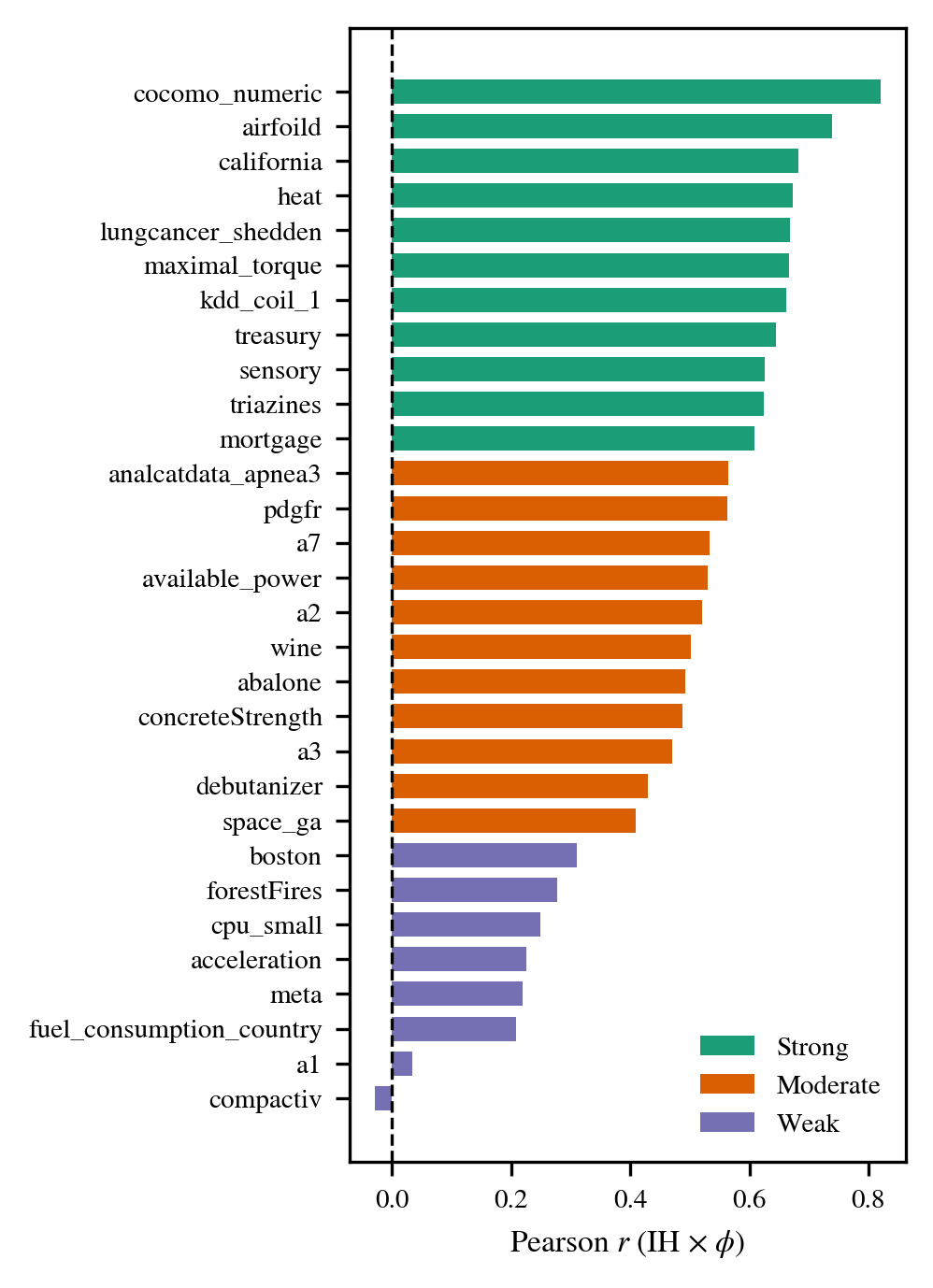}
\caption{Distribution of the correlation values between IH and $\phi$.}
\label{fig:histograma_correlacao_ih_phi}
\end{figure}

Figure~\ref{fig:histograma_correlacao_ih_phi} presents the distribution of Pearson correlation coefficients between the relevance function and the InHaR across the 29 evaluated datasets. The correlation values are grouped by absolute magnitude into three categories: strong ($|r| \geq 0.6$), moderate ($0.4 \leq |r| < 0.6$), and weak ($|r| < 0.4$). The predominance of moderate to strong correlations indicates that the two measures are related, but does not imply that they identify the same instances as relevant.

To further analyze this relationship at the instance level, instances were conceptually grouped into four categories based on whether they exhibit high or low values of $\phi$ and Instance Hardness. This analysis reveals instances that are statistically rare and difficult to predict (high $\phi$ / high IH), as well as those that are frequent and easy to learn (low $\phi$ / low IH). Importantly, a substantial proportion of instances fall into the high IH / low $\phi$ category, corresponding to observations that are not considered rare according to the relevance function, yet are consistently difficult to predict. Conversely, instances identified as rare by $\phi$ may exhibit low Instance Hardness, indicating that statistical rarity does not necessarily imply learning difficulty.

These results show that statistical rarity and prediction difficulty are not equivalent concepts. While IH and $\phi$ exhibit partial alignment, the InHaR captures difficult instances that are overlooked by relevance functions based solely on the target distribution, supporting its use in resampling.

\subsection{Applying InHaR to Bimodal Distributions}

Bimodal distributions of the target variable pose a well-known challenge for traditional relevance functions in imbalanced regression. In particular, relevance functions based solely on the statistical distribution of the target variable, such as the formulation proposed by Ribeiro~\cite{ribeiro2011utility}, assign fixed relevance values according to global distributional properties. As illustrated in Figure~\ref{fig:bimodal}, when the target distribution exhibits multiple modes, these functions tend to emphasize extreme values while failing to adequately capture low-density regions between modes, leading to an incomplete characterization of rarity~\cite{stocksieker2025comprehensive}.

To analyze this scenario in a controlled setting, we generated a synthetic regression dataset with a bimodal target distribution, as illustrated in Figure~\ref{fig:bimodal}. The dataset consists of two input features sampled from a standard normal distribution, while the target variable is generated from two distinct linear relationships with different offsets and coefficients. This process results in two well-separated modes in the target space, while maintaining overlapping regions in the input space, making the identification of rare or difficult regions non-trivial.

\begin{figure}[h]
\centering
\includegraphics[width=\columnwidth]{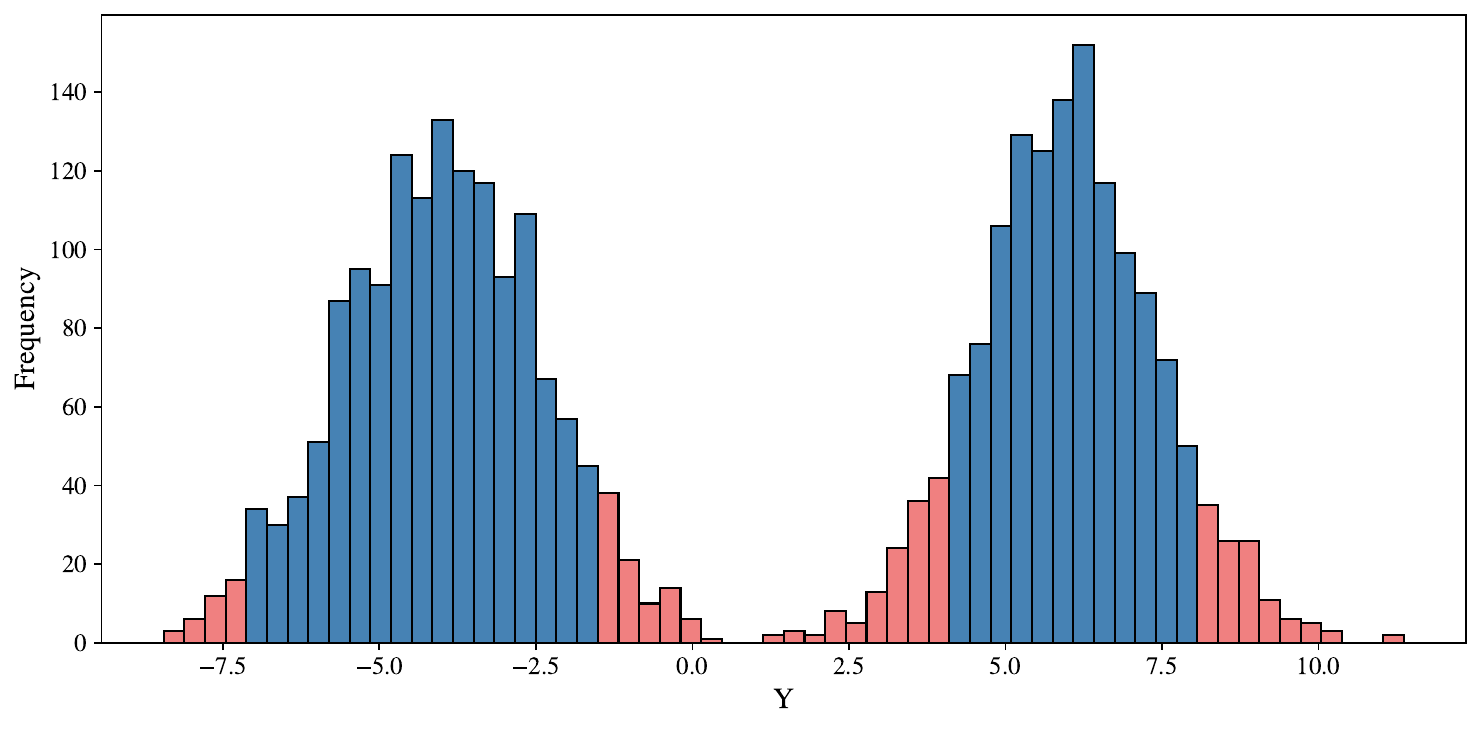}
\caption{Identification of rare regions in a bimodal target distribution using the median Instance Hardness (IH) per bin.}
\label{fig:ih_bimodal_success}
\end{figure}

Figure~\ref{fig:ih_bimodal_success} illustrates the behavior of the InHaR in the bimodal scenario. Unlike traditional relevance functions, the proposed approach identifies regions of the target distribution associated with higher prediction difficulty, even when these regions are not located at the extreme tails. Such regions correspond to portions of the bimodal structure that are harder to model and are therefore characterized as rare according to the InHaR criterion. This identification is performed by discretizing the target variable into equidistant bins and computing the median Instance Hardness value within each bin. A global threshold is then defined, and bins whose median IH falls within the top 30\% of observed values—corresponding to the 70th percentile of the IH distribution in our experiments—are labeled as rare. This percentile was selected based on preliminary experiments, while a more systematic sensitivity analysis is left for future work.
.

Following this analysis, experiments were conducted on this bimodal dataset to assess whether resampling strategies guided by Instance Hardness can improve predictive performance in regions that are systematically hard to learn. In this setting, traditional relevance functions were not considered, as they are unable to meaningfully define rarity in bimodal target distributions.

Table~\ref{tab:resultados_vitorias_bimodal} reports the comparative results obtained using InHaR-based resampling strategies within RO and GN. For both MAE and MSE, InHaR-based resampling achieved three wins out of the five evaluated models. These results indicate that prioritizing hard-to-predict instances during resampling can improve predictive performance even in scenarios where imbalance cannot be characterized solely by target frequency, reinforcing the role of Instance Hardness as an effective relevance criterion for complex target distributions.

\begin{table}[H]
\centering
\caption{Comparative results of wins in the bimodal distribution by method for MAE and MSE.}
\label{tab:resultados_vitorias_bimodal}
\setlength{\tabcolsep}{3pt}
\begin{tabular}{lccc|ccc}
\hline
\multirow{2}{*}{\textbf{Model}} 
& \multicolumn{3}{c}{\textbf{MAE}} 
& \multicolumn{3}{c}{\textbf{MSE}} \\
\cline{2-4}\cline{5-7}
& \textbf{None} & \textbf{InHaR-RO} & \textbf{InHaR-GN} 
& \textbf{None} & \textbf{InHaR-RO} & \textbf{InHaR-GN} \\
\hline
BG  & 0.0593 & \textbf{0.0592} & 0.1132 & \textbf{0.0124} & 0.0126 & 0.0326 \\
MLP & 0.0467 & \textbf{0.0423} & 0.0657 & 0.0046 & \textbf{0.0038} & 0.0082 \\
RF  & \textbf{0.0474} & 0.0480 & 0.1015 & 0.0103 & \textbf{0.0102} & 0.0286 \\
SVR & 0.1182 & 0.1154 & \textbf{0.0849} & 0.1155 & 0.1095 & \textbf{0.0508} \\
XGB & \textbf{0.0684} & 0.0691 & 0.1256 & \textbf{0.0106} & 0.0107 & 0.0326 \\
\hline
\end{tabular}
\end{table}

In a second analysis, the robustness of the proposed method was investigated through controlled variations in the synthetic data generation process, resulting in six distinct datasets. In all cases, the following characteristics were kept fixed: (i) 3,000 instances, (ii) two input variables, and (iii) a bimodal target distribution. The only factor varied was the level of Gaussian noise added to the data, ranging from 0 to 20.

\begin{figure}[H]
\centering
\includegraphics[width=\columnwidth]{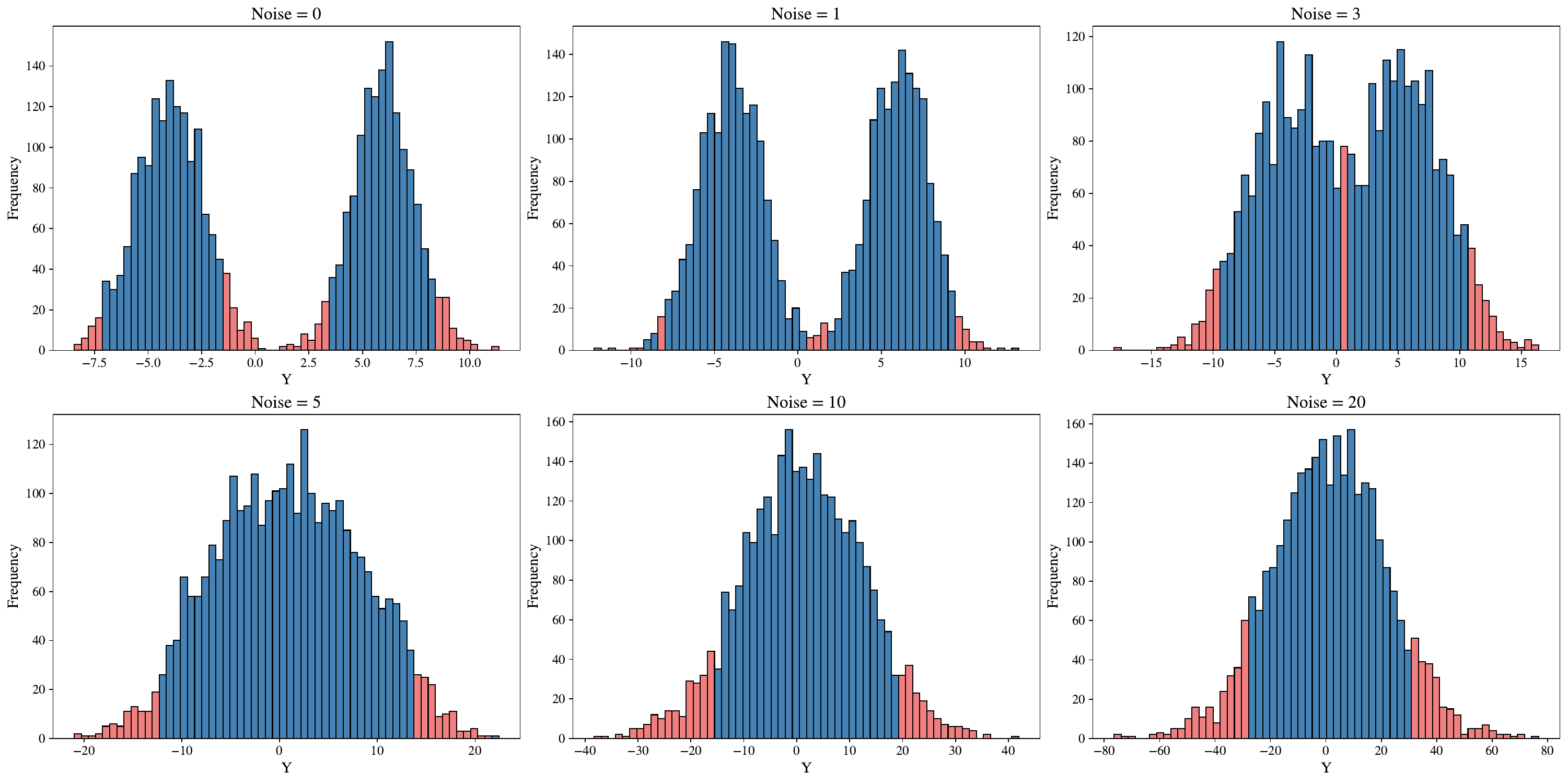}
\caption{Behavior of the InHaR under increasing noise levels in a bimodal target distribution.}
\label{fig:ih_var_ruido}
\end{figure}

Figure~\ref{fig:ih_var_ruido} illustrates the behavior of the InHaR under increasing noise levels. As noise increases, the target distribution gradually loses its clear bimodality and becomes smoother. Despite this change, the InHaR continues to highlight regions associated with higher prediction difficulty, particularly in less represented regions near the extremes of the target space. These results suggest that the proposed approach remains effective at identifying rare regions under moderate noise levels, even as the bimodal structure becomes less pronounced.

\subsection{Applying InHaR to Resampling Strategies}

\begin{table*}[t]
\centering
\caption{Comparative results of wins by method for MAE and MSE}
\label{tab:resultados_vitorias_geral}
\setlength{\tabcolsep}{3pt}
\begin{tabular}{lccccccc|ccccccc}
\hline
\multirow{2}{*}{\textbf{Model}} 
& \multicolumn{7}{c}{\textbf{MAE}} 
& \multicolumn{7}{c}{\textbf{MSE}} \\
\cline{2-8}\cline{9-15}
& \textbf{None} & \textbf{RO} & \textbf{InHaR-RO} & \textbf{DRF-RO} & \textbf{GN} & \textbf{InHaR-GN} & \textbf{DRF-GN}
& \textbf{None} & \textbf{RO} & \textbf{InHaR-RO} & \textbf{DRF-RO} & \textbf{GN} & \textbf{InHaR-GN} & \textbf{DRF-GN} \\
\hline
BG & 5 & 5 & \textbf{9} & 3 & 0 & 6 & 1 & \textbf{8} & 4 & 6 & 5 & 0 & 5 & 1 \\
MLP    & 4 & 6 & \textbf{10} & 4 & 1 & 4 & 0 & 7 & 5 & \textbf{8} & 4 & 1 & 4 & 0 \\
RF      & 6 & 6 & \textbf{8} & 2 & 0 & 6 & 1 & 6 & 5 & \textbf{9} & 4 & 0 & 5 & 0 \\
SVR     & \textbf{11} & 1 & \textbf{11} & 4 & 0 & 1 & 1 & 7 & 1 & 5 & \textbf{9} & 2 & 3 & 2 \\
XGB     & 6 & 6 & \textbf{10} & 4 & 0 & 3 & 0 & 6 & 6 & \textbf{10} & 4 & 0 & 3 & 0 \\
\hline
\textbf{Total} 
& 32 & 24 & \textbf{48} & 17 & 1 & 20 & 3 
& 34 & 21 & \textbf{38} & 26 & 3 & 20 & 3 \\
\hline
\end{tabular}
\end{table*}

In the resampling process, the InHaR is investigated as an alternative criterion for identifying relevant instances. To evaluate the generalization capability of the proposed methodology, an experimental analysis was conducted on 29 datasets (Table~\ref{tab:datasets_summary}), comparing MAE and MSE across different training strategies for regression models. Specifically, the following were considered: (i) the use of the original dataset, without applying resampling; (ii) resampling guided by the traditional relevance function ($\phi$) \cite{branco2016ubl}; (iii) resampling based on the \textit{Distance-Based Relevance Function} (DRF) \cite{inkim2025distance}, which takes into account the rarity of observations based on the distance between target values; and (iv) resampling guided by the InHaR, proposed in this work. 

Table~\ref{tab:resultados_vitorias_geral} presents an overview of the comparative results in terms of the number of wins achieved by each method for both MAE and MSE metrics. This table provides an initial descriptive analysis of the performance tendencies before considering statistical significance.

In this comparison, five distinct models were evaluated on 29 datasets, totaling 145 model--dataset pairs. As shown in Table~\ref{tab:resultados_vitorias_geral}, considering the MAE metric, the resampling functions that used the InHaR achieved superior performance in 68 pairs, with 48 associated with InHaR-RO and 20 with InHaR-GN. These results outperformed the other approaches in approximately 50\% of the evaluated pairs. For the MSE metric, 58 pairs performed better: 38 related to InHaR-RO and 20 to InHaR-GN, corresponding to about 40\% of the tested pairs.

Table~\ref{tab:wilcoxon_mae_mse} summarizes the pairwise Wilcoxon signed-rank test results, which were employed due to their suitability for paired comparisons when the normality of performance differences across datasets cannot be assumed. The results show a consistent advantage of the InHaR-based strategy across both RO and GN paradigms. For both MAE and MSE metrics, InHaR-based variants achieve substantially more wins over their respective baselines, with most of these improvements being statistically significant. Compared with the DRF variants, InHaR-based methods also show a higher number of wins, and a considerable portion of these differences is statistically significant. Importantly, when InHaR-based methods incur losses, these are predominantly non-significant, indicating that the proposed strategy rarely leads to statistically meaningful performance degradation. Overall, the results confirm that the InHaR function provides robust and statistically reliable improvements for both RO and GN.

\begin{table}[H]
\centering
\caption{Wilcoxon pairwise comparison (InHaR vs. others). Wins/losses and significant results ($p<0.05$) for MAE and MSE.}
\label{tab:wilcoxon_mae_mse}
\setlength{\tabcolsep}{3pt}
\begin{tabular}{lcc|cc}
\hline
\multirow{2}{*}{Comparison} &
\multicolumn{2}{c}{MAE} &
\multicolumn{2}{c}{MSE} \\ \cline{2-5}
 & Win (sig.) & Loss (sig.) & Win (sig.) & Loss (sig.) \\ \hline
InHaR-RO vs RO     & \textbf{27 (22)} & 2 (0)  & \textbf{20 (13)} & 9 (1)  \\
InHaR-RO vs DRF-RO & \textbf{17 (11)} & 12 (0) & \textbf{13 (9)}  & 16 (1) \\
InHaR-GN vs GN     & \textbf{24 (19)} & 5 (4)  & \textbf{20 (19)} & 9 (5)  \\
InHaR-GN vs DRF-GN & \textbf{23 (15)} & 6 (3)  & \textbf{16 (8)}  & 13 (8) \\ \hline
\end{tabular}
\end{table}

\subsubsection{Performance Analysis on Rare Instances}

Model performance is evaluated on subsets of rare instances identified under different relevance criteria. Mean Absolute Error (MAE) and Mean Squared Error (MSE) are computed over instances classified as rare either by the traditional relevance function ($\phi$) or by the proposed InHaR, with the goal of analyzing how resampling strategies behave when assessed on the regions prioritized by each criterion.

Table~\ref{tab:rare_subset_comparison} presents pairwise Wilcoxon comparisons between InHaR-based resampling methods and their respective baselines (RO and GN), considering only instances labeled as rare under each relevance definition. Results are reported as wins and losses across datasets, with statistically significant differences ($p<0.05$) indicated in parentheses.

\begin{table}[t]
\centering
\caption{Pairwise comparison of InHaR-based resampling methods against RO and GN on subsets of instances defined by different relevance criteria. Wins and losses across datasets are reported, with statistically significant results ($p<0.05$) shown in parentheses.}
\label{tab:rare_subset_comparison}
\setlength{\tabcolsep}{3pt}
\begin{tabular}{llcc}
\hline
\textbf{Subset definition} & \textbf{Metric} & \textbf{InHaR-RO vs RO} & \textbf{InHaR-GN vs GN} \\
\hline
$\phi$-rare & MAE & 9 (7) / 20 (20) & 7 (6) / 22 (22) \\
$\phi$-rare & MSE & 9 (7) / 20 (20) & 7 (6) / 22 (22) \\
\hline
InHaR-rare & MAE & \textbf{15 (9)} / 14 (6) & \textbf{16 (9)} / 13 (11) \\
InHaR-rare & MSE & \textbf{15 (9)} / 14 (6) & \textbf{16 (9)} / 13 (11) \\
\hline
Normal ($\phi$ \& InHaR) & MAE & \textbf{26 (26)} / 3 (2) & \textbf{27 (26)} / 2 (2) \\
Normal ($\phi$ \& InHaR) & MSE & \textbf{26 (26)} / 3 (2) & \textbf{27 (26)} / 2 (2) \\
\hline
\end{tabular}
\end{table}

When evaluation is restricted to instances identified as rare by the traditional relevance function, InHaR-based resampling methods exhibit fewer wins than losses across both MAE and MSE. This behavior is expected, as resampling is guided by a different relevance criterion. In contrast, when evaluation focuses on instances identified as rare by the InHaR, resampling guided by Instance Hardness consistently achieves more wins across both RO and GN for both error metrics. These results indicate that relevance-based resampling strategies should be evaluated within the regions defined by their own relevance criteria, as assessing one criterion on subsets identified by another may lead to misleading conclusions.

An additional analysis was conducted on instances classified as normal by both relevance criteria. In this scenario, InHaR-based resampling methods exhibit a clear advantage over standard RO and GN, with a large number of statistically significant wins and very few losses. This behavior indicates that, unlike traditional relevance-based resampling, the InHaR does not introduce substantial performance degradation in well-represented and easier regions of the data space.

These results highlight an important trade-off in relevance-guided resampling. Traditional relevance functions tend to aggressively oversample statistically rare regions, often at the expense of performance in normal regions. By contrast, the InHaR prioritizes instances based on learning difficulty, resulting in a less aggressive and more balanced resampling strategy that improves performance on difficult cases while preserving accuracy on normal instances.

\section{Conclusion}

In conclusion, the results show that defining the InHaR function is a viable alternative to traditional distribution-based relevance functions, particularly in bimodal target distributions where static relevance assignments fail to capture local variations in learning difficulty. Using relevance-independent metrics (MAE and MSE), resampling strategies guided by the InHaR consistently improved predictive performance when compared to both the original data and traditional relevance-based resampling. The use of relevance-independent metrics avoids potential bias that may arise when evaluation measures are directly coupled with the relevance definition under analysis, while still indicating that prioritizing hard-to-predict instances can effectively guide learning.

Finally, it is necessary to identify the limitations of the proposed approach. Since it relies on a pool of machine learning models, the computational cost may be high for datasets with a large number of records. Moreover, the choice of the threshold plays an important role in controlling how many instances are prioritized during resampling, potentially affecting model performance. While a fixed threshold was adopted based on preliminary experiments, a systematic analysis of threshold sensitivity was not conducted, leaving open the question of whether a single threshold range is sufficient across datasets or whether dataset-specific tuning is required.

\bibliographystyle{IEEEtran}
\bibliography{bib}

@article{avelino2024resampling,
  title   = {Resampling strategies for imbalanced regression: a survey and empirical analysis},
  author  = {Avelino, J. G. and Cavalcanti, G. D. C. and Cruz, R. M. O.},
  journal = {Artificial Intelligence Review},
  volume  = {57},
  number  = {82},
  year    = {2024}
}

@techreport{torquette2024instancehardness,
  title       = {Instance hardness measures for classification and regression problems},
  author      = {Torquette, G. P. and Nunes, V. S. and Paiva, P. Y. A. and Lorena, A. C.},
  institution = {Instituto Tecnológico de Aeronáutica (ITA)},
  address     = {São José dos Campos, SP, Brazil},
  year        = {2024},
  note        = {Published: 27 February 2024}
}

@article{inkim2025distance,
  title   = {Distance-Based Relevance Function for Imbalanced Regression},
  author  = {In, D. D. and Kim, H.},
  journal = {Stats},
  volume  = {8},
  number  = {3},
  pages   = {53},
  year    = {2025},
  doi     = {10.3390/stats8030053}
}

@misc{stocksieker2025comprehensive,
  title  = {A Comprehensive Survey on Imbalanced Regression: Definitions, Solutions, and Future Directions},
  author = {Stocksieker, S. and Pommeret, D.},
  year   = {2025},
  note   = {HAL open archive, hal-05213741}
}

@article{kowatschImbalanceRegression,
  title={Imbalance in regression datasets},
  author={Kowatsch, Daniel and M{\"u}ller, Nicolas M and Tscharke, Kilian and Sperl, Philip and B{\"o}tinger, Konstantin},
  journal={arXiv preprint arXiv:2402.11963},
  year={2024}
}

@article{paiva2022instancehardness,
  title   = {Relating instance hardness to classification performance in a dataset: a visual approach},
  author  = {Paiva, P. Y. A. and Moreno, C. C. and Smith-Miles, K. and Valeriano, M. G. and Lorena, A. C.},
  journal = {Machine Learning},
  volume  = {111},
  number  = {8},
  pages   = {3085--3123},
  year    = {2022}
}

@article{ribeiro2020imbalanced,
  title   = {Imbalanced regression and extreme value prediction},
  author  = {Ribeiro, R. P. and Moniz, N.},
  journal = {Machine Learning},
  volume  = {109},
  pages   = {1803--1835},
  year    = {2020},
  doi     = {10.1007/s10994-020-05900-9}
}

@incollection{torgo2009utility,
  title     = {Utility-Based Regression},
  author    = {Torgo, L. and Ribeiro, R. P. and Pfahringer, B.},
  booktitle = {Lecture Notes in Computer Science},
  volume    = {5812},
  publisher = {Springer},
  year      = {2009}
}

@article{haixiang2017learning,
  title   = {Learning from class-imbalanced data: Review of methods and applications},
  author  = {Guo, Haixiang and Li, Yijing and Shang, Jennifer and Gu, Mingyun and Huang, Yuanyue and Gong, Bing},
  journal = {Expert Systems with Applications},
  volume  = {73},
  pages   = {220--239},
  year    = {2017}
}

@article{krawczyk2016learning,
  title={Learning from imbalanced data: open challenges and future directions},
  author={Krawczyk, Bartosz},
  journal={Progress in Artificial Intelligence},
  volume={5},
  number={4},
  pages={221--232},
  year={2016},
  publisher={Springer}
}

@article{johnson2019survey,
  title={Survey on deep learning with class imbalance},
  author={Johnson, Justin M and Khoshgoftaar, Taghi M},
  journal={Journal of Big Data},
  volume={6},
  number={1},
  pages={1--54},
  year={2019},
  publisher={Springer}
}

@article{branco2016ubl,
  title={UBL: an R package for utility-based learning},
  author={Branco, Paula and Ribeiro, Rita P and Torgo, Luis},
  journal={arXiv preprint arXiv:1604.08079},
  year={2016}
}

@article{branco2019pre,
  title={Pre-processing approaches for imbalanced distributions in regression},
  author={Branco, Paula and Torgo, Luis and Ribeiro, Rita P},
  journal={Neurocomputing},
  volume={343},
  pages={76--99},
  year={2019},
  publisher={Elsevier}
}

@article{ribeiro2011utility,
  title={Utility-based regression},
  author={Ribeiro, R},
  journal={Ph. D. dissertation},
  year={2011},
  publisher={Dep. Computer Science, Faculty of Sciences-University of Porto}
}

@inproceedings{torgo2009precision,
  title={Precision and recall for regression},
  author={Torgo, Luis and Ribeiro, Rita},
  booktitle={International Conference on Discovery Science},
  pages={332--346},
  year={2009},
  organization={Springer}
}

@inproceedings{torgo2013smote,
  title={Smote for regression},
  author={Torgo, Lu{\'\i}s and Ribeiro, Rita P and Pfahringer, Bernhard and Branco, Paula},
  booktitle={Portuguese conference on artificial intelligence},
  pages={378--389},
  year={2013},
  organization={Springer}
}

@inproceedings{branco2017smogn,
  title={SMOGN: a pre-processing approach for imbalanced regression},
  author={Branco, Paula Oliveira and Torgo, Lu{\'\i}s and Ribeiro, Rita Paula},
  booktitle={First International Workshop on Learning with Imbalanced Domains: Theory and Applications}, 
  volume={74},
  pages={36--50},
  year={2017}
}

@article{dougherty1989nonnegativity,
  title={Nonnegativity-, monotonicity-, or convexity-preserving cubic and quintic Hermite interpolation},
  author={Dougherty, Randall L and Edelman, Alan S and Hyman, James M},
  journal={Mathematics of Computation},
  volume={52},
  number={186},
  pages={471--494},
  year={1989}
}

@misc{tukey1970exploratory,
  title={Exploratory Data Analysis, limited prelim. ed},
  author={Tukey, JW},
  year={1970},
  publisher={Addison-Wesley, Reading, Mass}
}

@article{camacho2022geometric,
  title={Geometric SMOTE for regression},
  author={Camacho, Lu{\'\i}s and Douzas, Georgios and Bacao, Fernando},
  journal={Expert Systems with Applications},
  pages={116387},
  year={2022},
  publisher={Elsevier}
}

\end{document}